# Novel and Automatic Parking Inventory System Based on Pattern Recognition and Directional Chain Code

Reza Azad*,Majid Nazari**
*Shahid Rajaee Teacher Training University, Tehran , Iran, rezazad68@gmail.com
** Shahid Rajaee Teacher Training University, Tehran , Iran, doliskani@gmail.com

***Abstract:*** *The objective of this paper is to design an efficient vehicle license plate recognition System and to implement it for automatic parking inventory system. The system detects the vehicle first and then captures the image of the front view of the vehicle. Vehicle license plate is localized and characters are segmented. For finding the place of plate, a novel and real time method is expressed. A new and robust technique based on directional chain code is used for character recognition. The resulting vehicle number is then compared with the available database of all the vehicles so as to come up with information about the vehicle type and to charge entrance cost accordingly. The system is then allowed to open parking barrier for the vehicle and generate entrance cost receipt. The vehicle information (such as entrance time, date, and cost amount) is also stored in the database to maintain the record. The hardware and software integrated system is implemented and a working prototype model is developed. Under the available database, the average accuracy of locating vehicle license plate obtained 100%. Using 70% samples of character for training, we tested our scheme on whole samples and obtained 100% correct recognition rate. Further we tested our character recognition stage on Persian vehicle data set and we achieved 99% correct recognition.*

**1.Introduction**

Vehicle License Plate Recognition (VLPR) also known as Automatic Vehicle License Plate Recognition (AVLPR) was invented in 1976. Many scientist groups took interest in VLPR after 1990s with the development of digital camera and the increase in processing speed. VLPR is an image processing technology which enables to extract vehicle license number form digital images. It consists of a still or video camera which takes the image of vehicle, find the location of the number in the image and then segments the characters and by using the k -Nearest Neighbour (KNN) scheme, it translates the license number of pixel value into numerical or string. VLPR can be used in many areas such as parking inventory [1], [2], security control of restricted areas [3], traffic law enforcement [4]-[6], congestion pricing [7], and automatic toll collection [8].

Typical VLPR System consists of four modules: image acquisition, license plate extraction, character segmentation, and character recognition [9]. The efficiency and accuracy of the system largely depends on the second module and various approaches have been used for this purpose. To detect the region of car license plate, many techniques have been used. The algorithm presented in [10] using projection and Euclidean distance reaches 87% as its performance. The algorithm presented in [11] using sliding concentric windows and probabilistic NN reaches 86% as its overall performance. What is reported in [12] using filtering and template matching depicted its performance as 91%. The method presented in [13] with Gabor filter and connected component reported its plate detection rate as 91.7%. The algorithm reported in [14] applying edge analysis and feed forward NN reaches 92.3% as its character recognition rate.

In [15] adaptive boosting (AdaBoost) is combined with Haar-like features to obtain cascade classifiers for license plate extraction. The Haar-like features are commonly used for object detection. Using the Haar-like features makes the classifier invariant to the brightness, colour, size, and position of license plates. In [16], a new and fast vertical edge detection algorithm (VEDA) was proposed for license plate extraction. VEDA showed that it is faster than Sobel operator by about seven to nine times. In [17] and [18] combination of edge statistics and mathematical morphology showed very good results, but it is time consuming and because of this problem, [19] uses block-base algorithm.

In [20] a novel method called "N row distance" is implemented. This method scans an image with N row distance and counts the existent edges. If the number of the edges is greater than a threshold then the license plate is recognized, if not threshold have to be reduced and algorithm will be repeated. This method is fast and has good results for simple images. Disadvantage of this paper is that the edge based algorithms are sensitive to unwanted edges such as noise edges, and they fail when they are applied to complex images.

A wavelet transform-based algorithm is used in [21] for extraction of the important features to be used for license plate location. This method can locate more than one license plate in an image. Methods which are symmetry based are mentioned in [22]. In [23], firstly, it takes the input image into a grayscale, then for analyzing the location of plate the operation of morphology such as erosion and dilation is applied, and the plate is extracted with use of vertical and horizontal projection among various candidates. In [24] the plate is a location with the black background and white writings. In this way that, firstly, takes the image into the HSI and applies the capability of being black colour of its background for this purpose, it uses a mask and segments the image according to HSI colour intensity parameter and creates a binary image. For cancelling probable noises, it uses the operation of erosion and dilation, then labels the existing candidates and for cancelling the candidates which aren't the location of plate, it applies the geometric capability of the plate and other characters, then for recognizing a primary candidate, it uses the colour intensity histogram, and recognizes the location of plate.

The current paper aims at investigation into and identification of the novel Iranian plate characterized by both inclusion of blue area on it and its geometric shape. Obviously, the suggested system contains suitable velocity due to not making use of heavy pre-processing operation such as image-improving filters, edge-detection operation and omission of noise at the beginning stages. So, the recommended method of ours is compatible with model-adaptation, i.e., the very blue section of the plate so that the present method indicated the fact that if several plates are included in the image, the method can successfully manage to detect it. We implemented this work for parking inventory System. The proposed VLPR system is efficient and novel so that it can run real time using normal desktop PC and can recognize various standard number plates such as public in yellow background, the government in red background and private cars in white background, under acceptable lighting conditions.

The proposed system consists of five steps. (1) Detection of the vehicle and capturing the image of front view of vehicle. (2) Extract and localizing number plate with use of pattern recognition. (3) Number plate segmentation and character separation. (4) K -Nearest Neighbour classifier is used to convert characters of pixel value to alphanumeric value. (5) Using detected license number to charge entrance cost accordingly, storing details in database, receipt generating and communicating with hardware to automate parking barriers. This research work is ordered as follow; section 2 presents the proposed vehicle plate location and recognition system, Section 3 briefly discusses the hardware and database part to make an efficient automated parking inventory system, Sections 4 and 5 describe experimental results, and conclusion respectively.

**2.Proposed Method**

Out proposed method for VLPR, consisted of 3 stages which in the first stage, the entire candidate of plate places are exploited, in the second stage, the real place of plate is determined among exploited candidates with use of plate geometry and characters color jumping and finally in the third stage extracted characters are recognized. General diagram of theproposed method for VLPR is shown in Fig. 1. Stage1 and stage2 are detailed in [25].

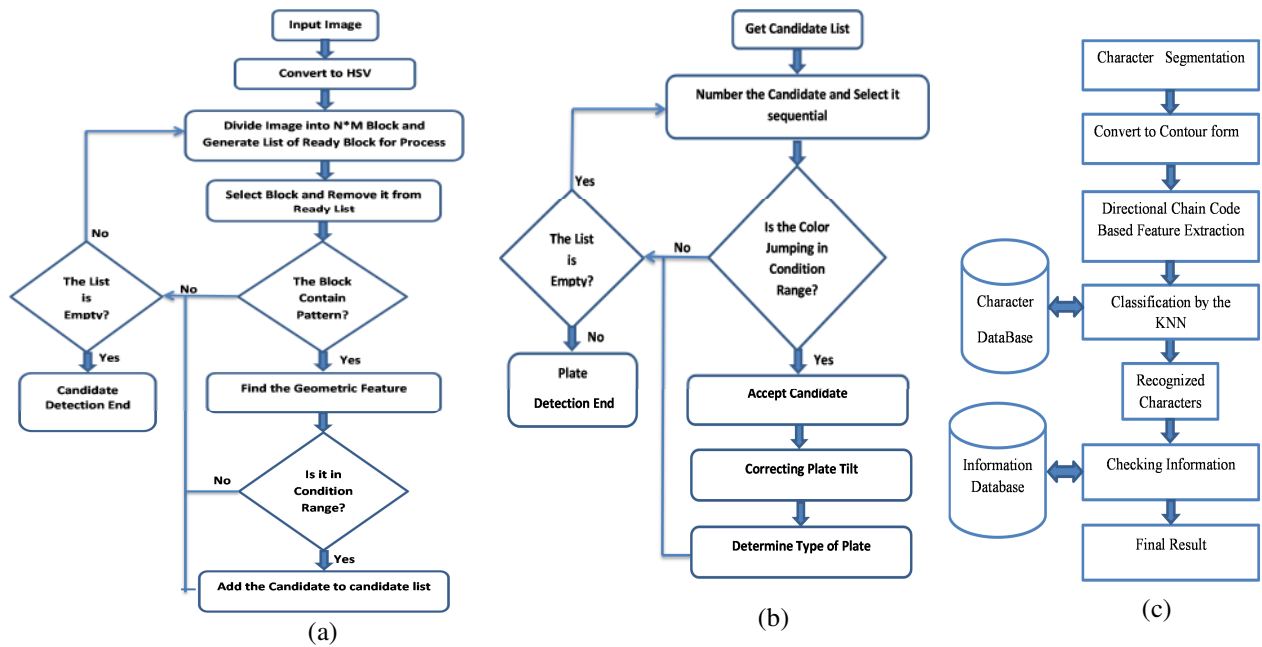

Fig. 1: General diagram of the proposed method for VLPR (a): Stage one (b): Stage two (c): stage three

## 2.1 Character Segmentation and Recognition (stage three)

To isolate the characters of car license plate, many techniques have been used. In [26], the extracted license plate is resized into a known template size. In this template, all character positions are known. After resizing, the same positions are extracted to be the characters. This method has the advantage of simplicity. However, in the case of any shift in the extracted license plate, the extraction results in background instead of characters. Since characters and license plate backgrounds have different colours, they have opposite binary values in the binary image. Therefore, some proposed methods as in [27]-[38] project the binary extracted license plate vertically to determine the starting and the ending positions of the characters, and then project the extracted characters horizontally to extract each character alone. Segmentation is performed in [39]–[44] by labelling the connected pixels in the binary license plate image. The labelled pixels are analysed and those which have the same size and aspect ratio of the characters are considered as license plate characters. This method fails to extract all the characters when there are joined or broken characters.

In this paper from last step, the area that is exploited as a plate, first probable noising are solved, then plate image is complemented till its writing of plate inside is seen such white violence. Then this area is labelled and through the available regions, the regions that are bigger are stored as exploited characters in 30*15 sizes. Fig. 2 shows histograms of extracted palte after removing noise witches the most space are relevant characters.

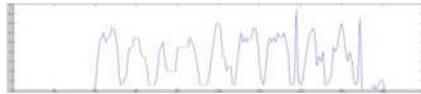

Fig. 2: Histogram of extracted plate

### 2.1.1 Character Recognition

The extracted characters are then recognized and the output is the license plate number. To recognition the characters of car license plate, many techniques have been used. In [45], the feature vector is generated by dividing the binary character into blocks of 3×3 pixels. Then, the number of black pixels in each block is counted. In [46], the feature vector is generated by dividing the binary character after a thinning operation into $3 \times 3$ blocks and counting the number of elements that have 0°, 45°, 90°, and 135° inclination. In [47], the character is scanned along a central axis. This central axis is the connection between the upper bound horizontal central moment and lower bound horizontal central moment. Then the number of transitions from character to background and spacing between them form a feature vector for each character. This method is invariant to the rotation of the character because the same feature vector is generated. Template matching is performed in [48] - [50] after resizing the extracted character into the same size. Several similarity measuring techniques are defined in the literature. Some of them are Mahalanobis distance and the Bayes decision technique [48], Jaccard value [49], Hausdorff distance [50].

For recognition of characters in this paper we used directional chain code. Directional chain code information of the contour points of the input image can be used as features for different purposes like character segmentation, recognition, etc. [52]. In our system we computed features based on chain-code directional frequencies of contour pixels of the images as follows: First we found the bounding box (minimum rectangle containing the character) of each input image which is a two-tone image. Then for better result and independency of features to size, font and position (invariant to scale and translation), we converted each image to a normal size of 30×15 pixels. Then For each image the chain code frequencies for all 8 directions were computed. By this action 120features will be produced. We considered these feature as our feature set. In Fig. 3(a), a normalized image with its bounding box is shown. We extracted the contour of the normalized image Fig. 3(b).

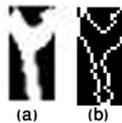

Fig. 3: (a): Bounding box of extracted character   (b): character in contour form

### 2.1.2. Classification

Classification stage uses the features extracted in the feature extraction stage for deciding the class membership. Classification phase is the decision making phase of a VLPR engine. In this work, we have used k-NN classifier for recognition. In the k-nearest neighbour classifier, Euclidean distances from the candidate vector to stored vector are computed. The Euclidean distance between a candidate vector and a stored vector is given by equation (1),

$$d = \sqrt{\sum_{k=1}^{N}(x_k - y_k)^2} \quad (1)$$

Here, N is the total number of features in feature set, $x_k$ is the library stored feature value and $y_k$ is the candidate feature value.

### 2.2. Determine Type of Plate

For specifying the kind of Iran country plates that are in three categories: 1) the public in yellow background, 2) the government in red background and 3) private cars in white background, are known. The below algorithm is used, that it recognize the most of the colour frequencies as a plate kind.

---

Function Determining type of plate (plate)
(1): [n, m] ←size (plate)
(2): (∀i, j: i, j ∈ plate) Do {3, 4}
(3): X= $\begin{cases} \text{Red:} & \text{if } \{S \geq 0.45 \ And \ V \geq 0.5 \ And \ 0.8 \leq H \leq 0.94\} \\ \text{Yellow:} & \text{if } \{S \geq 0.45 \ And \ V \geq 0.5 \ And \ 0.58 \leq H \leq 0.74\} \\ \text{White:} & \text{f } \{S \leq 0.15 \qquad And \ V \geq 0.8\} \end{cases}$
(4): Increase the Class(X) one time
(5): Kind=max (Class)
Show ("Type of Plate Color is (Kind)");
   End Function

---

## 3. Hardware Model

The VLPR system is interfaced with hardware model and database to make an automated parking inventory system. The hardware model consists of proximity sensor to detect the presence of vehicle, a web camera to capture the image, motors to open/close the parking barriers, desktop computer on which VLPR algorithm is executed, LCD, seven segments display and a microcontroller for controlling all the components of hardware model. As the vehicle arrives at parking, the inductive proximity sensor detects the vehicle and gives a signal to the PC using parallel port. The camera connected to the PC captures the image of front view of the vehicle and applies VLPR algorithm on the image to recognize the vehicle's license plate number. This number is then used to charge entrance cost and generate receipt containing all the information of vehicle. Also, all the information such as time, date, plate number and cost amount is stored in database to maintain the record. PC then sends the signal to microcontroller using parallel port and the road barrier is opened for a time by driving motors and "You Have Permission to go" is displayed on the LCD to guide the vehicles. Complete hardware design of the system is shown in Fig. 4.

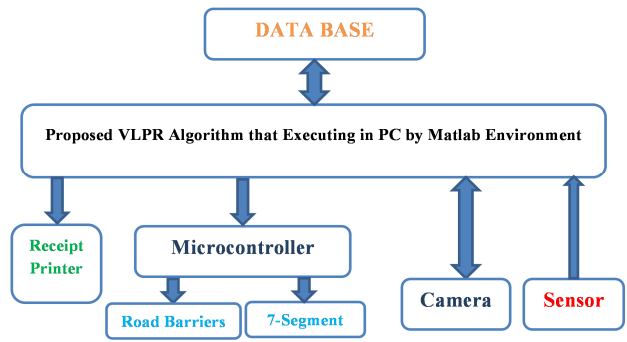

Fig. 4: The system hardware design

## 4. Practical Result

Our suggestive method have been done on Intel Core i3-2330M CPU, 2.20 GHz with 2 GB RAM under Matlab 2011 environment. According to the previous license plate recognition (LPR) approaches, there is not a popular database of LPR to evaluate the performance of its methods especially for our method that need front side view images. In this respect, in order to prove the quality of proposed approach, 100 images are captured by a 2 megapixel camera of mobile phone (Nokia 5230) from front view of car. These images are 640*480 sizes, RGB mode, variety in point of view, various light condition, various distances and backgrounds. Next, the proposed approach is applied on them, and the accuracy rate is computed. Table 1 shows the result.
0

TABLE I: Performance of the Proposed Method

| Total image of car | Total characters | Correct license plate location | Correct character recognition | Percent Efficiency for each part | |
|---|---|---|---|---|---|
| 100 | 800 | 100 | 800 | 100% | 100% |

Further In our research experiment, we used [53] dataset that contains numerical and character images and all those image samples are extracted from the real world environment, such as parking lot, freeway and etc. for showing high accuracy of our method. In the table 2, our character recognition method is compared with normal factoring method represented in [25] and template matching mentioned in [53] that both of them used this dataset for evaluation of their works.

TABLE 2: Character recognition accuracy

| Total Image | Method | Technique | Correct character recognition | Percent Efficiency |
|---|---|---|---|---|
| **1200** | S.H.M Kasaei and etc. [54] | Template Maching | 1104 | 92% |
| | R.Azad and etc.[25] | Normal Factoring | 1164 | 97% |
| | R.Azad, H.R. Shayegh [Proposed Method] | Direction Chain Code | 1188 | 99% |

The suggestive method is robustness against parameters such as: the different size and situation of plate in image, the view of videotaping and different light situation in videotaping, injuries and pollution of plate. In suggestive method when vehicle plate has tilt, the system is able to identify and solve it that the success rate of this operator has been gained to 100%.In general, the advantages that are presented in this paper compared to other methods include the following: Lower computational complexity; Fast response and operation; Ability to correct plate tilt; Ability to implementation on microprocessors; Usability in real time work; Detect minimum candidates as car plates; In most images, the proposed method detects a candidate that is in fact an original license plate; Scale invariant; Ability to detect multi plates in an image; Detecting all plate types of Iran e.g. public plates with yellow background and governmental plates with red background; Recognizing private, governmental and public plates; Ability to use as an automatic parking inventory system; Robust method for character recognition.

5. Conclusion

In this paper, we proposed a novel and efficient method for Vehicle License Plate Recognition and implementation of that method for automatic parking inventory system. The system has been tested on many images of various lighting conditions and system can be implemented on parking entrance for managing parking. In the proposed method for finding the place of plate the new technique based of pattern recognition is used that the system is worked quite well. For recognition of character new technique based on directional chain code suitably are used that showed high accuracy in character recognition stage. In suggestive system if the plate had a tilt its tilt removed away and recognized correctly. The suggestive methods that expressed in this paper, is tested under available data set that the success rate of plate location and character recognition rate achieved 100%. This state shows the high efficiency rate of top expressed method.